\documentclass{statsoc}
\usepackage[letterpaper, margin = 1in]{geometry}
\usepackage{url}
\usepackage{graphicx}
\usepackage[caption=false]{subfig}
\usepackage{amsmath}
\usepackage{amssymb}
\usepackage{natbib}
\usepackage{hyperref}
\usepackage{epsfig}
\usepackage{arydshln}
\usepackage{multirow}
\usepackage{epsfig}
\usepackage{appendix}
\usepackage[table]{xcolor}
\usepackage[normalem]{ulem}

\newcommand{\xVec}{{\bf x}}

\title[Ensembles of phalanxes across assessment metrics]{Ensembles of phalanxes across assessment metrics for robust ranking of homologous proteins}

\author{Jabed H.\ Tomal}
       \address{Department of Mathematics and Statistics,
       Thompson Rivers University,
       Kamloops, BC V2C 0C8, Canada.}
			\email{jtomal@tru.ca}
\author{William J.\ Welch}
       \address{Department of Statistics,
       The University of British Columbia,
       Vancouver, BC V6T 1Z4, Canada.}
       \email{will@stat.ubc.ca}
\author[Tomal, Welch, and Zamar]{Ruben H.\ Zamar}
       \address{Department of Statistics,
       The University of British Columbia,
       Vancouver, BC V6T 1Z4, Canada.}
       \email{ruben@stat.ubc.ca}

\begin{document}
\begin{abstract}

Two proteins are homologous if they have a common evolutionary origin,
and the binary classification problem is to identify proteins in a candidate set
that are homologous to a particular native protein.
The feature (explanatory) variables available for classification are various 
measures of similarity of proteins.
There are multiple classification problems of this type for different native proteins and
their respective candidate sets.
Homologous proteins are rare in a single candidate set,
giving a highly unbalanced two-class problem.
The goal is to rank proteins in a candidate set according to the probability
of being homologous to the set's native protein.  
An ideal classifier will place all the homologous proteins
at the head of such a list.  
Our approach uses an ensemble of models in a classifier and an ensemble of assessment metrics. For a given metric a classifier combines models, each based on a subset of the available feature variables which we call phalanxes. 
The proposed ensemble of phalanxes identifies strong and diverse subsets of feature variables. A second phase of ensembling aggregates classifiers based on diverse evaluation metrics. The overall result is called an ensemble of phalanxes and metrics.
It provide robustness against both \emph{close} and \emph{distant} homologues.

\end{abstract}

\section{Introduction}\label{sec:intro}

Identifying rare-class items in a highly unbalanced two-class dataset 
occurs in many applications. Examples include 
identifying fraudulent activities in credit card transactions
\citep{Bolton:2002,Bhusari:2011},
intrusion detection in internet surveillance \citep{Lippmann:2000,Sherif:2002,Kemmerer:2005,Maiti:2012,Gendreau:2016},
detection of terrorist attacks or threats \citep{Fienberg:2005,Peeters:2006,Cohen:2014},
finding information on the world wide web \citep{Gordon:1999,Nachmias:2002,Al-Masri:2008,Baeza-Yates:2011},
and finding drug-candidate biomolecules in a chemical library \citep{Bleicher:2003,WangHou:2010,Tomal:2016}.

The application in this paper concerns homology of proteins. 
Two or more proteins are homologous if they share a common evolutionary origin or ancestry. 
Understanding of homology status helps scientists to infer evolutionary sequences of proteins \citep{Koonin:2003,Lorenza:2009}. 
It is also important in bioinformatics \citep{Soding:2005,Seung-Zin:2016},
with widespread applications in prediction of a protein's function, 3D structure, and
evolution \citep{Bork:1998,Henn:2001,Kinch:2003,Debora:2011,Meier:2015,Waterhouse:2018}. 

Evolutionarily related homologous proteins have similar amino-acid sequences and 3D structures. 
Hence, feature (explanatory) variables $x_1, \ldots, x_d$ for classification of 
homology status are based on similarity scores
between a candidate protein and the native (target) one
 \citep{Vallat:2009}.
A few more details will be given in Section~\ref{sect:data}.

In this article, then, the classification problem is as follows.
Let $Y$ be a random variable representing the homology status of a protein, taking value $1$
for the rare homologous class and $0$ otherwise. 
Let  $\xVec = (x_1, \ldots, x_d)$ be the available similarity-score features 
for estimating the probability $\pi$  that  $Y=1$, that is,  
$\pi = E(Y \mid \xVec)$. 
We build our model using training data for which $y$ and $\xVec$ are both known. 
The trained model is then used to predict the unknown class status $y$ of proteins in a test dataset for which only $\xVec$ is known. 
Going beyond such a standard set up,
there are actually many such classification problems, one for each of many native proteins.
In the training data a native protein has an associated block of data of proteins 
with known homology status. 
As homologous proteins are usually rare in the candidate set for a single native protein,
we combine training data across all native proteins when building a classifier.
The classifier is then tested on a new set of native proteins,
each with its own block of proteins to classify.
How to assess accuracy on the test set is an important aspect of the proposed methodology, described in Section~\ref{sect:evmetrics}.


Several methods have already been applied in protein homology prediction,
such as hidden Markov models \citep{Karplus:1998},
support vector machines (SVMs) and neural networks \citep{Hochreiter:2007}, and Markov random fields \citep{Ma:2014}.
The cited methods are based directly on one trained model only, i.e., they are not ensembles of classifiers.

Ensemble methods that combine multiple models are generally considered more powerful for prediction \citep[e.g.,][]{Dietterich:2000}.
For example, the overall winner in the protein homology section of the $2004$ KDD competition was the Weka Group \citep{Weka:2004}. 
(The competition's collection of datasets is used in this article.) 
They tried a large number of algorithms and selected the top three classifiers: 
$(1)$ a boosting ensemble \citep{Freund:1997} of $10$ unpruned decision trees,
$(2)$ a linear SVM \citep{Cortes:1995} with a logistic model fitted to the raw outputs for improved probabilities,
and $(3)$ $10^5$ or more random rules, a variant of RF \citep{Breiman:2001}. The top three classifiers,
where two are already ensemble methods, were aggregated to obtain the winning ensemble.

Several groups have published methodology for the $2004$ KDD Cup protein homology data since the competition. \cite{Bachem:2015} proposed a Bayesian nonparametric model to cluster proteins using Dirichlet Process (DP)-Means, which is similar to K-Means clustering. They used coresets, a data summarization method from computational geometry, with DP-Means to cluster proteins. \cite{Tsai:2016} compared the performances of distributed computing and cloud computing methods based on MapReduce in terms of computational time and classification accuracy. They used support vector machines as a tool to mine information from big data without any intention of developing a statistical model or ensemble. We note they used 10-fold cross-validation (CV) to assess classification accuracy, rather than the provided test data (for which we will provide results). 


It is well recognised that ensembling is most effective when diverse models are combined.
Popular methods for creating diversity are random perturbation
of the training data \citep{Melville:2003,Breiman:1996,Breiman:2001},
possibly combined with random perturbation or clustering of the feature variables 
\citep{Breiman:2001,Wang:2005,Gupta:2014,TomWelZam:2015,Tomal:2016,TomWelZam2017},
or random projections of subsets of the feature variables \citep{CanSam2017}.
In contrast, we are unaware of previous work on injecting diversity into ensembles through a variety of evaluation metrics.

We propose methods that seek diversity by creating ensembles at two levels.
First, for a given metric, following \cite{TomWelZam:2015, Tomal:2016} with important adaptations
we construct an ensemble of phalanxes (EPX).
Phalanxes are disjoint subsets of features chosen in a data-adaptive way using the metric in an objective function.
One classifier is fit to each phalanx,
and their various estimates of the probability of homology are averaged for an EPX classifier.
Second, different metrics in phalanx construction lead to different EPX models;
their probability estimates are again averaged in an ensemble of metrics (EOM).
We call the overall ensemble of phalanxes and metrics EPX-EOM.

When the class of interest is rare, 
metrics of prediction accuracy need to go beyond minimizing misclassification error. 
A typical block of about $1000$ proteins will have approximately $5$ homologous to the native protein. 
A naive classifier that calls all the candidates ``non-homologous'' would therefore achieve a $0.5\%$ misclassification rate or 99.5\% accuracy when in practice it has no power 
to identify homology.
Instead, we evaluate a classifier by its ability to rank candidate proteins 
such that the rare homologous cases appear at the top of a ranked list. 
The candidate with the largest probability of homology has rank 1, 
followed by the protein with the second largest probability and rank 2, etc.
There are several ways to assess the quality of a ranked list,
as described in Section~\ref{sect:evmetrics} along with some comments on how to deal with 
tied probabilities,
leading to different metrics for building models and evaluating them.

The different metrics focus on diverse aspects of a ranked list.
Some are heavily focused on performance at the beginning of the list:
are the candidates with the largest probabilities of homology actually homologous?
Success at the beginning of the list is more likely for homologues that
are close to the native protein as measured by the feature variables
and are hence easier to classify.
In contrast, another metric---called rank last---records the position in the list of the last
true homologue.
Some true homologues are difficult to classify correctly, 
and hence appear well down the ranked list,
because they are ``remote'' with respect to the block's native protein.
There is interest in detecting them 
to infer function of uncharacterized proteins and improve genomic annotations \citep{Eddy:1998,Eddy:2011,Soding:2005,Kaushik:2016}.
Optimization of rank last helps achieve this latter goal. 
Thus, aggregating EPX models across the diverse metrics aims to achieve good performance for all homologues,
whether they are close or remote.
In contrast, methods competing in the KDD Cup competition were optimized and assessed
for one ranking-performance metric at a time.
The results in Section~\ref{sec:results} show that it is possible to find 
a single robust ensemble that performs as well according to all measures.

The rest of the article is organized as follows. Section \ref{sect:data} describes the protein homology data and the feature variables available for classification.
Section \ref{sect:evmetrics} defines three evaluation metrics specific to the ranking application and illustrates how their diverse characteristics are potentially helpful for the application. Section~\ref{sect:ensemble} describes the algorithm for construction of an ensemble classifier:
how to find phalanxes of variables to serve in base logistic-regression models
and how both phalanxes and complementary evaluation metrics are used in an overall ranking ensemble.
Section \ref{sec:results} showcases the results of the ensemble and makes comparisons with
the winners of the $2004$ KDD Cup competition and other state-of-the-art ensembles. 
Finally, Section \ref{dis:con} summarizes the results and draws conclusion.


\section{Protein homology data} \label{sect:data}

The protein homology data are downloaded from the $2004$ knowledge discovery and data mining (KDD) cup competition website\footnotemark[1].
Registration is required to download the data and submit
predictions for test proteins where only the feature variables are known. 
The site reports back for the test data the performance measures 
we describe in Section~\ref{sect:evmetrics}.
\footnotetext[1]{\url{http://osmot.cs.cornell.edu/kddcup/datasets.html}: Accessed June 20, 2019.}

The data are organized in blocks, where the proteins in a block relate to one native protein.
In total, there are $303$ blocks randomly divided into training and test sets,
with $153$ and $150$ native proteins, respectively.
The block identifiers are not used for model building: models are built across all the
training data.
The identifiers are required, however, to compute the assessment metrics described in Section \ref{sect:evmetrics},
which are averaged over blocks.
The minimum, first quartile, median, third quartile and maximum block sizes are
$612$, $859$, $962$, $1048$ and $1244$, respectively, in the training set and 
$251$, $847$, $954$, $1034$ and $1232$ in the test set.
The distributions of block sizes are therefore similar except that the test set contains $3$
small blocks of sizes $251$, $256$ and $372$. 

The response variable $y \in \{0, 1\}$ is $1$ if a candidate protein is homologous 
to the block's native protein, and $0$ otherwise.
The homology status is known for the $145,751$ proteins in the training set
and unknown for the $139,658$ in the test set. 
The training set contains relatively few homologous proteins.
The minimum, first quartile, median, third quartile and maximum of the within-block proportions of homologous proteins
are $0.00080$, $0.00143$, $0.00470$, $0.01206$ and $0.05807$, respectively. 
More than $75\%$ of the training blocks contain at most $2$ homologous proteins per $100$ candidates. This highly unbalanced classification problem requires appropriate evaluation
metrics, as considered in Section \ref{sect:evmetrics}.

\begin{figure}
        \centering
        \subfloat[]{
                \includegraphics[width=0.47\textwidth]{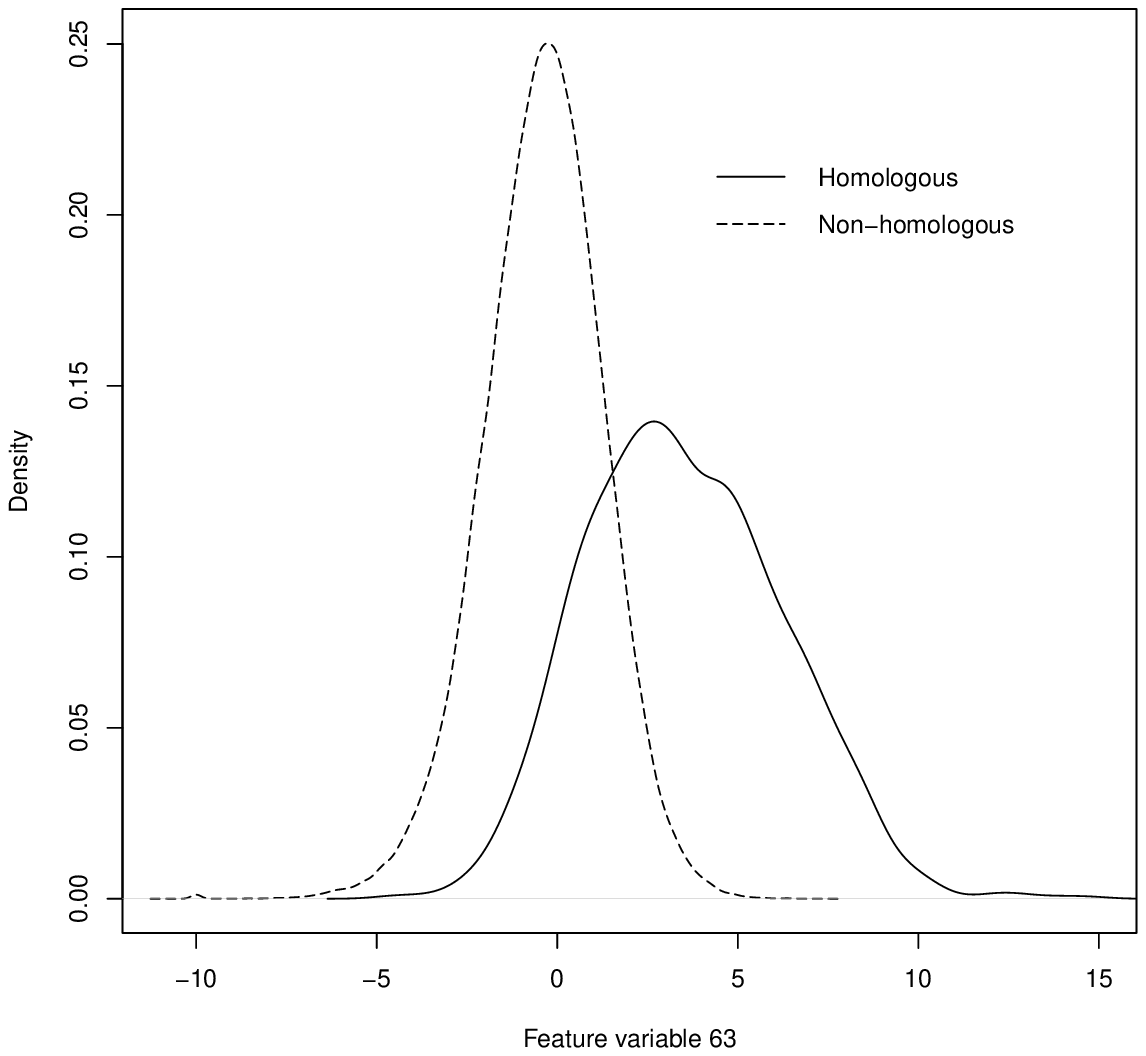}
                \label{feature:63}}
        \subfloat[]{
                \includegraphics[width=0.47\textwidth]{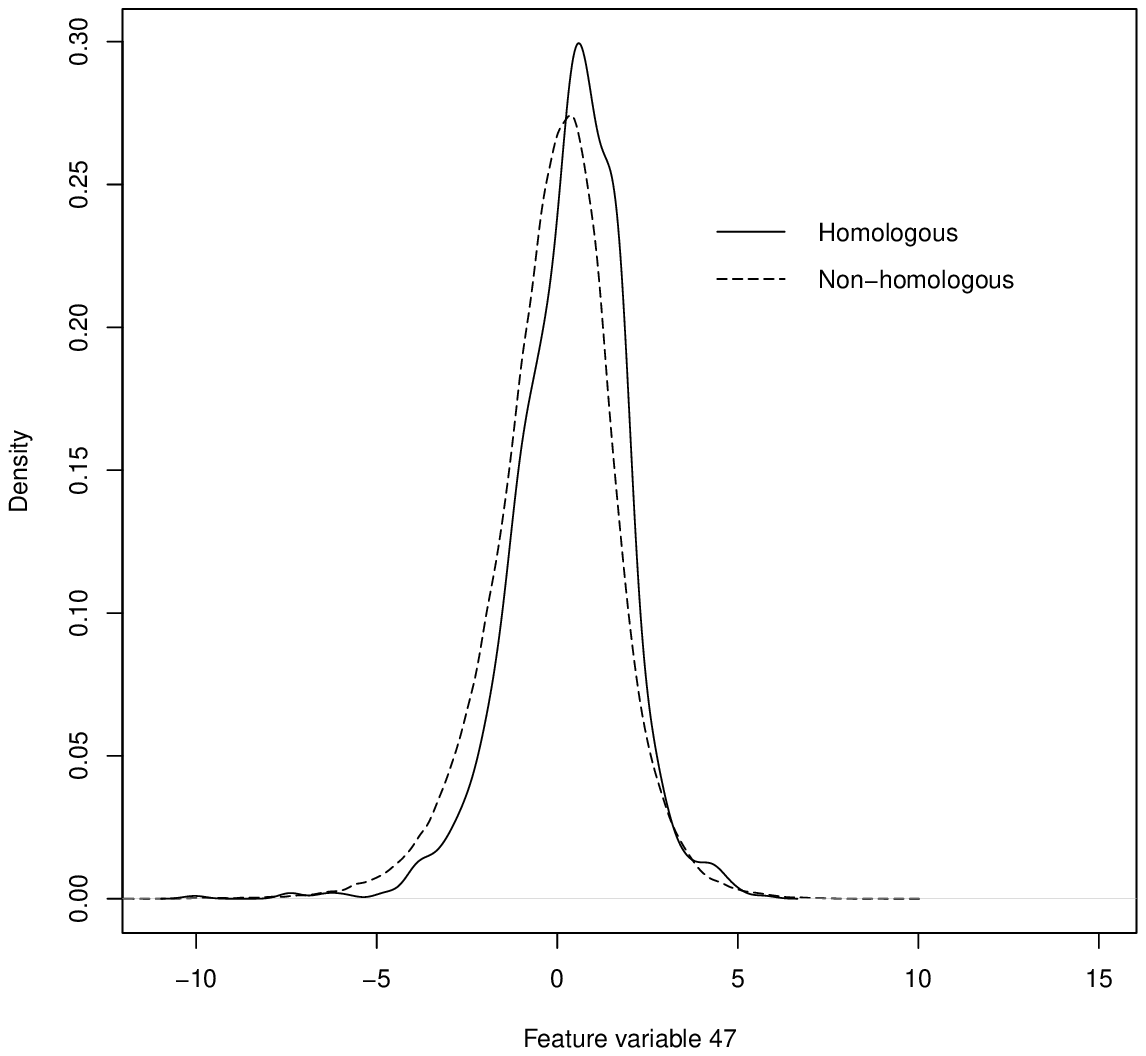}
                \label{feature:47}}
\caption{Density plots of the feature variables $63$ (panel a) and $47$ (panel b) for the homologous (solid line) and non-homologous proteins (dashed line) in the training data.}
\label{density_plots}
\end{figure}

The 74 feature variables available in the training and test sets for classification 
represent structural similarity 
and amino acid sequence identity
between a native protein and a candidate protein.
The variables include measures of global and local sequence alignment, global and  local structural fold similarity (protein threading),
position-specific scoring and profile (PSI-BLAST), 
and so forth.
The appendix of \cite{Vallat:2008} contains further details.
Some of the feature variables are clearly potentially helpful for classification.
For instance, Figure \ref{density_plots}(a) shows density plots of 
feature variable $63$ in the training data, conditioned on homology status.
This variable seems to differentiate the two classes fairly well.
In contrast, 
Figure \ref{density_plots}(b) shows analogous density plots for feature variable $47$,
where the two classes are not well separated.
The presence of apparently  less-informative feature variables motivates us to perform variable selection.

\section{Assessment metrics and diversity} \label{sect:evmetrics}

As discussed in Section~\ref{sec:intro}, performance will be assessed in terms
of how well a method's estimated probabilities of homology for a list of candidates
place true homologues ahead of non-homologues.
After providing some visual intuition, 
we describe three metrics to measure specific properties of a ranking, to be used
for training an algorithm and for assessment

\subsection{Hit curves} \label{sec:hit:curve}

\begin{figure}
	        \centering
	        \subfloat[]{
	        	\includegraphics[width=0.47\textwidth]{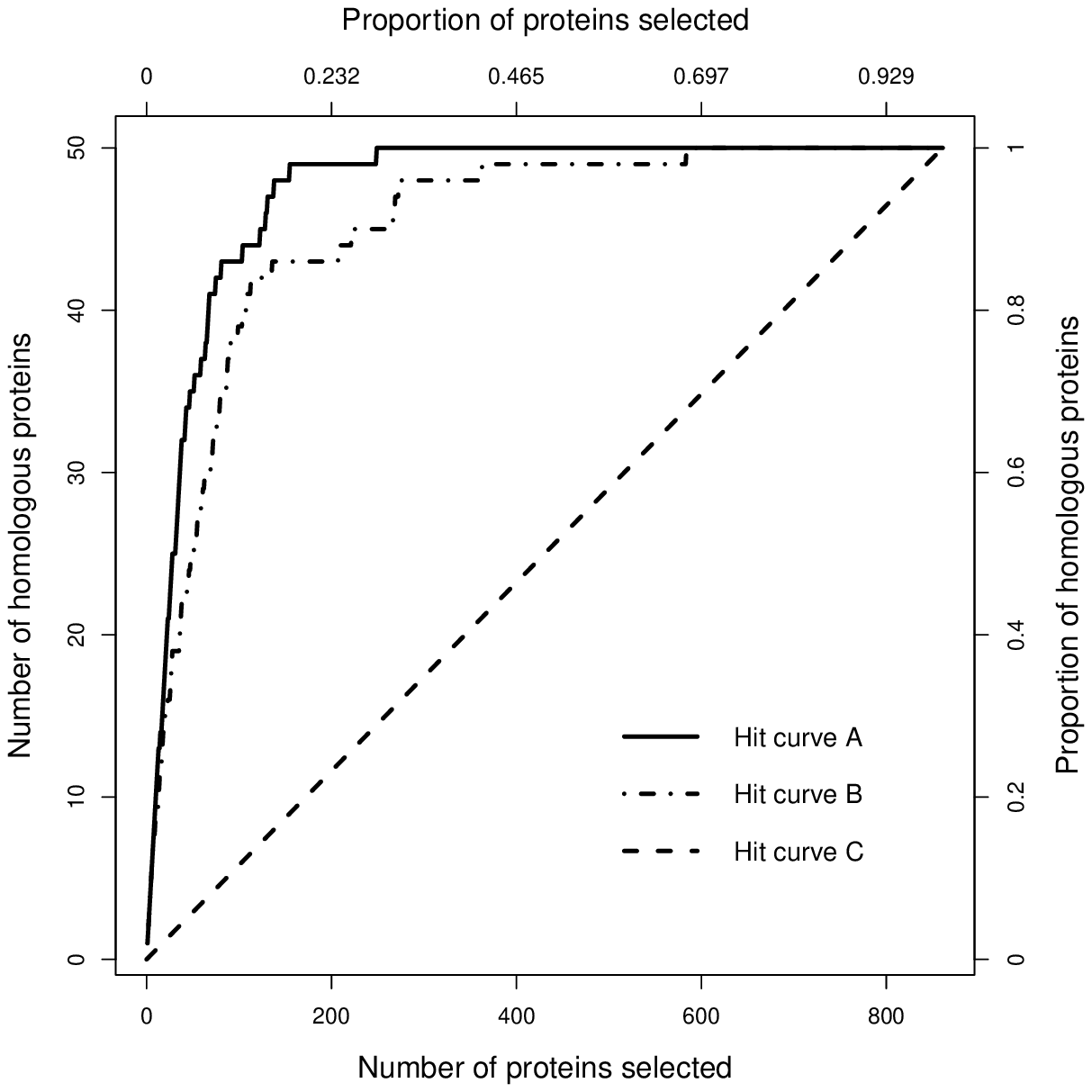}
	        	\label{sub:fig:example1}}
	        \subfloat[]{
	        	\includegraphics[width=0.47\textwidth]{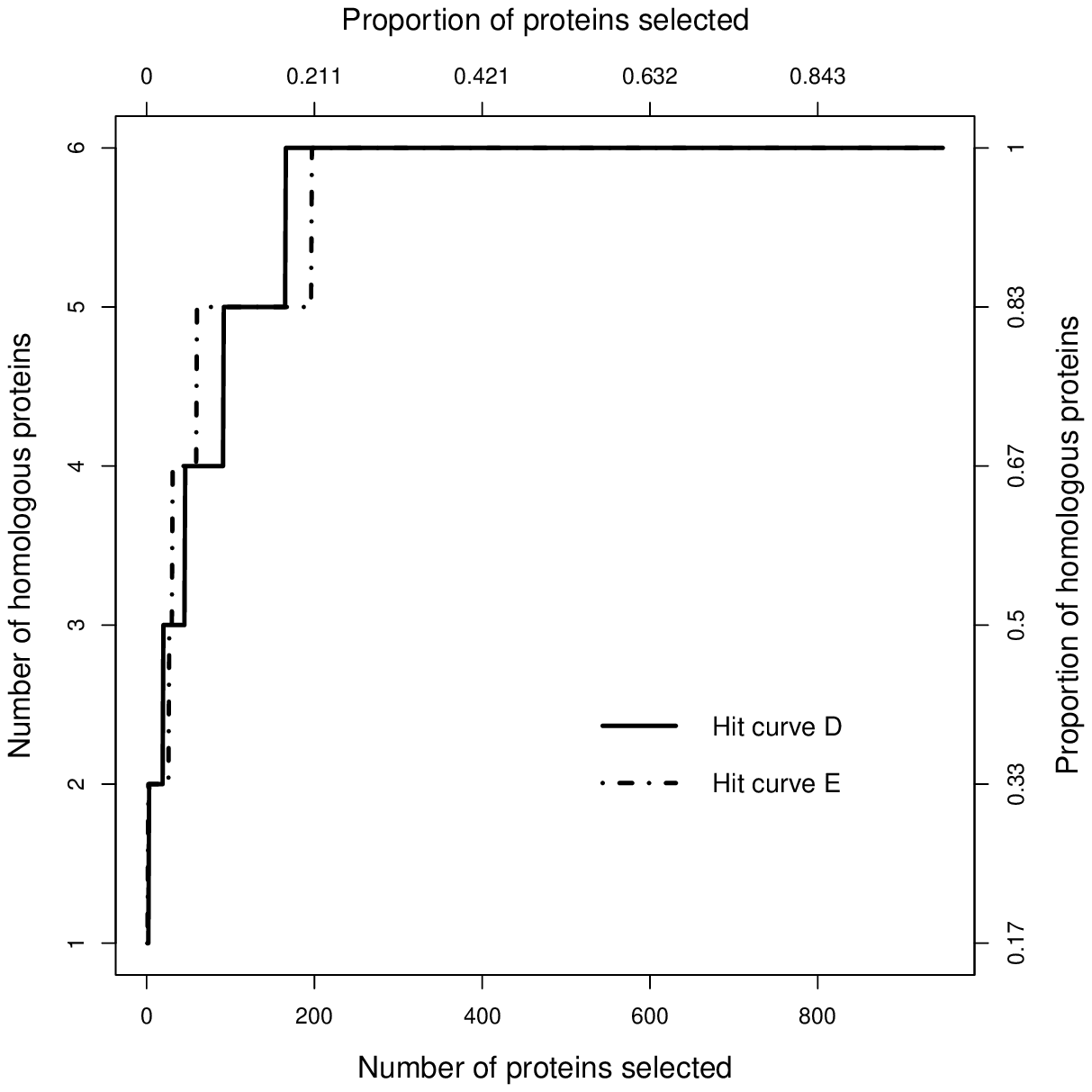}
	        	\label{sub:fig:example2}}
	\caption{Example hit curves. Panel (a) shows curves $A$, $B$ and $C$. Hit curve $A$ is superior to $B$, while $C$ has performance only comparable with random ranking. In panel (b) hit curves $D$ and $E$ cross each other at several locations making their comparison more difficult.}
	\label{fig:hitcurveintro}
\end{figure}

Overall ranking performance can be visualised by a hit curve. 
Suppose the $n$ cases in a test set are ranked by some method to have nonincreasing estimated probability of homology.
Out of $h$ homologous proteins to be found,
a shortlist of the top $t \in \{1, \ldots, n\}$ ranked proteins has 
$h_t \in \{0, 1, 2, \dots, \min(h, t)\}$ homologous proteins or ``hits''. 
The plot of $h_t$ against $t$ is called a hit curve.
Figure \ref{fig:hitcurveintro}(a) illustrates three hit curves for a particular candidate set (block ID $238$) with $n = 861$ proteins, of which 50 are homologous.
Hit curve $A$ is uniformly better than hit curve $B$, as $h^{A}_{t} \geq h^{B}_{t}$
for every $t \in \{1, \ldots, n\}$. On the other hand, a classifier with a hit curve close to the diagonal line $C$ shows performance similar to random ranking.
Ideally, we want hit curves that demonstrate an initial maximal slope of 1 hit per candidate until all the hits are exhausted.

Hit curves are difficult to optimize numerically as they are difficult to characterize by a unique scalar objective. Sometimes they cross each other, making comparison even more difficult. 
For instance, Figure \ref{fig:hitcurveintro}(b) shows curves $D$ and $E$
for a block (block ID $254$) of candidates with $n = 949$ and $h = 6$.
The curves cross at several points, and hence neither is uniformly superior. 
Using several numerical summaries of a hit curve provides an opportunity to ensemble classification models trained for different desirable aspects of a curve.


Three metrics are used to assess
the ranking performance of a model. Since the data come
in blocks, each of the three metrics is computed in each block separately. Given a
metric, the average performance across the blocks is used as the final value of the metric.
Thus, in order to perform well a model has to rank the rare homologous proteins well
across many blocks. The definitions of the three assessment metrics are given below.

\subsection{Average precision} \label{sec:apr}

Average precision (APR)
is a variant of the metric average hit rate \citep{MuZhu:2004}, and its application is common in information retrieval \citep{Baeza-Yates:2011}.
From the $h_t$ of a hit curve in Section~\ref{sec:hit:curve}, define
\[
\ a(t) =\frac{h_t}{t} \in [0,1],
\]
which is the precision (or \emph{hit rate}) computed at $t$.
Let $1 \leq t_1 < t_2 < \dots < t_h \leq n$ be the positions in the ranked list of the $h$ homologous proteins. Then APR is defined as 
\begin{equation} \label{eqn:apr}
\text{APR} = \frac{1}{h} \left[a(t_1) + a(t_2) + \dots + a(t_h) \right].
\end{equation}
It averages ``precisions'' at the ranking positions of the homologous proteins.
A larger APR value implies better predictive ranking, with the maximum $1$ reached if and only if all of the homologous proteins appear at the beginning of the list, i.e., $h_t = t$ for $t = 1, \ldots, h$ and hence all $a(t_j) = 1$ in (\ref{eqn:apr}). 
APR is highly influenced by performance at the beginning of the list.
For instance, if the first hit is found in the second versus first positions, its
contribution to APR drops from $1/h$ to $1 / (2h)$.


\subsection{Rank last} \label{sec:rkl}

As its name implies, the metric rank last (RKL) is simply the position of the last homologous protein in the ranked list.  Hence, it is a measure of worst-case classification accuracy, and smaller values are better.

Ties are treated conservatively: if the last homologous protein is in a group of proteins---homologous or not---with tied probabilities of homology, the last member of the group determines RKL.
Thus, RKL is always at least $h$,
and the maximum possible value is the size of the block. 

\subsection{TOP1} \label{sec:top1}

If the top ranked candidate is homologous to the native protein,
TOP1 scores $1$ for success, otherwise it is $0$.
TOP1 is calculated conservatively when there are ties: if the largest estimated probability of homology occurs multiple times,
all the candidates in the group must be homologous to score $1$; 
any non-homologues lead to a TOP1 score  of $0$. Thus, again, it is never beneficial to have tied probabilities. The goal is to maximize TOP1.


\subsection{Diversity in assessment metrics}\label{sec:diversity:metrics}

The following toy example illustrates how the three assessment metrics are complementary.
Table \ref{compl:apr:rkl} shows APR, RKL, and TOP1 from five classifiers
ranking two homologues in a large number of candidates. 
\begin{table}
\caption{\label{compl:apr:rkl}A toy example showing complementary behaviour of the metrics.}
\centering
\fbox{%
\begin{tabular}{||c|rl|lrc||}
  \hline\hline
& \multicolumn{2}{l|}{Ranks of the} & \multicolumn{3}{c||}{Metric}\\
\cline{4-6}
Classifier &  \multicolumn{2}{l|}{homologues} & APR & RKL & TOP1 \\\hline
1 &  1, &  2   & 1 & 2 & 1 \\
2 &  1, & 20   & 0.55 & 20 & 1 \\
3 &  2, &  4   & 0.50 & 4 & 0 \\\hline
4 &  8, & 200 & 0.0675 & 200 & 0 \\
5 & 10, & 50 & 0.0070 & 50 & 0 \\
 \hline\hline
\end{tabular}}
\end{table}
Classifier 1 ranks both homologous proteins at the head of the list,
and is therefore optimal according to all metrics.
Classifiers 2 and 3 have similar APR scores, because classifier 2 obtains a large APR contribution from the homologue ranked 1,
which more than offsets the relatively poor rank of 20 for the second homologue.
Similarly, TOP1 favours classifier 2 over 3, 
but RKL gives a large penalty to classifier 2 because the second homologue
is not found until position 20.
Rows $4$ and $5$ show much weaker classifiers (or the results for a difficult block).  
TOP1 cannot separate these classifiers, and they have similarly poor APR scores.
Their RKL values better reflect the weaker worst-case performance of classifier 4, however, which does not find the second homologue until position 200 in the ranked list. 
We also note from this example that APR and TOP1 both give considerable weight to the head of the list, but the very discrete TOP1 fails to separate the fairly good performance of classifier 3 from the much poorer performances of classifiers 4 and 5.


\section{Ensembling across phalanxes and assessment metrics} \label{sect:ensemble}

\subsection{Algorithm of phalanx formation}
For a fixed assessment metric, e.g., APR, the EPX algorithm generates a set of phalanxes, where the variables in a phalanx will give one prediction model, and those models are ensembled.  
The method for hierarchical clustering of variables into phalanxes draws from \cite{TomWelZam:2015} with the following important differences.
\begin{enumerate}
    \item The base classifier is logistic regression instead of random forests (RF).
    \item Whereas RF allows estimation of an assessment metric via  out-of-bag samples, with logistic regression we need to use CV
    (10-fold for all computations).
    \item There is no initial grouping of variables.
    \item The criterion for merging two groups of variables into one is modified.
    \item The criterion for filtering out a candidate phalanx is modified.
\end{enumerate}
In addition, after  building one EPX classifier based on APR and and another using RKL, the two ensembles will be further ensembled.
While the details of the algorithm are reported in the appendix, we next provide rationale for these modifications.

\subsubsection{Base classifier: logistic regression} \label{lrmodel}



The EPX algorithm chooses subsets of variables to be used by a base classifier.  Thus there is flexibility to choose a base classifier that performs well for the given application.
A submission to the KDD Cup by the research group \cite{Ruben-Will:2004}
predicted the same  protein-homology data fairly well using logistic regression (LR). Some new preliminary comparisons building EPX models (optimized for APR) with LR as the base method versus RF confirmed the superior performance of LR.  The earlier work applied variable selection to logistic regression, which is not necessary here as the step-wise construction of candidate phalanxes and their final screening provides implicit variable selection.
Another advantage of logistic regression is that it 
is computationally much cheaper than RF and many other classification methods. As well as improving predictive performance with reduced computational burden,
the use of LR demonstrates that EPX is adaptable to a base learner other than 
the previously used RF.

\subsubsection{Cross-validation instead of out-of-bag}

To evaluate predictive performance of an LR model,
throughout we use 10-fold CV at the block level.
For each block the randomly chosen indices defining its 10 folds
are saved and used to evaluate all models during phalanx formation.
Note that the use of CV is necessary here as RF's out-of-bag option
does not apply to LR. 

\subsubsection{No initial grouping of variables}

Previous versions of the EPX algorithm were applied mainly to large sets of
binary predictor variables, which were grouped at the outset for two reasons.
First, a single binary variable will usually give a very weak classifier.
Second the computational complexity of EPX is such that running time
increases with the square of the number of initial variables;
reducing 1000s of binary variables to a smaller number of initial groups made computation feasible.
In contrast, the variables for the protein-homology application are not binary (see Figure \ref{density_plots}) and the dimensionality is modest---74 variables. Hence, this step is unnecessary.

\subsubsection{Criterion for merging groups of variables} \label{sect:phalanx}

Let $a$ be an assessment metric (larger the better for definiteness), and at any iteration of phalanx formation let
$\mathbf{g}_i$ for $i = 1, \ldots, s$ be a set of groups of the predictor variables. 
Initially, each $\mathbf{g}_i$ contains one of the available feature variable.

Consider two distinct groups $\mathbf{g}_i$ and $\mathbf{g}_j$; 
their two LR models have metric values $a_i$ and $a_j$, respectively.
If the groups are merged, their single LR model gives $a_{ij}$ for the metric, 
whereas ensembling their two respective LR models gives $a_{\overline{ij}}$. 
To avoid degrading the performance of a group of important variables by merging
it with another containing less useful 
variables, our criterion for merging combines the two groups that minimize
\[
\ m_{ij} = \frac{\max(a_{\overline{ij}}, a_i, a_j)}{a_{ij}}
\]
over all possible pairs $i$ and $j$. 
If $a_{ij} > \max(a_{\overline{ij}}, a_i, a_j)$,
the performance of the new single model using the union of the variables is better than
their ensemble performance, i.e., 
\begin{equation} \label{eqn:joint:ensemble}
a_{ij} > a_{\overline{ij}}
\end{equation}
and better than the individual performances, i.e.,
\begin{equation} \label{eqn:joint:individual}
a_{ij} > a_i \quad \text{and} \quad a_{ij} > a_j .
\end{equation}
When both (\ref{eqn:joint:ensemble}) and (\ref{eqn:joint:individual}) hold, $m_{ij}$ is less than $1$
and the groups of variables $\mathbf{g}_i$ and $\mathbf{g}_j$ are merged together.
After each merge,
the number of groups $s$ is reduced by $1$, and one of the new groups is the union of two old groups. 
The algorithm of \cite{TomWelZam:2015} minimizes $m_{ij} = a_{\overline{ij}}/a_{ij}$, thus taking account of (\ref{eqn:joint:ensemble}) but not (\ref{eqn:joint:individual}).

The algorithm continues until
$m_{ij} \geq 1$ for all $i, j$, suggesting that merging either degrades individual performances or ensembling performance.
At completion, $s$ and the $\mathbf{g}_i$ define $c = s$ candidate phalanxes 
of variables $\mathbf{x}_i = \mathbf{g}_i$ for $i = 1, \ldots, c$, to pass to the next step.

\subsubsection{Criterion for filtering the candidate phalanxes}

Our algorithm identifies candidate phalanxes that help all other phalanxes in the final ensemble; other candidate phalanxes are filtered out.

To detect weak and/or harmful phalanxes, the
following criterion is minimized
\[
\ f_{ij} = \frac{a_{\overline{ij}}}{\max(a_i, a_j)}
\]
over all possible pairs of candidate phalanxes $(\mathbf{x}_i$ and $\mathbf{x}_j)$.
If $a_{\overline{ij}} < \max(a_i, a_j)$, the ensembling performance of candidate phalanxes $\mathbf{x}_i$ and $\mathbf{x}_j$
is weaker than that of the better performing individual phalanx. Equality of $a_{\overline{ij}}$ and $\max(a_i, a_j)$ shows that the ensemble
of the two does not improve upon individual performances. In these cases, 
$f_{ij} \leq 1$ and the weaker phalanx is filtered out.
After filtering out a harmful phalanx, the number of candidate phalanxes $c$ is reduced by $1$, and the algorithm iterates until $f_{ij} > 1$ or $c = 1$. 
Such filtering always retains the strongest candidate phalanx, and
all $p$ phalanxes in the final set $\{\mathbf{x}_{(1)}, \mathbf{x}_{(2)}, \ldots, \mathbf{x}_{(p)}\}$ help overall ensembling performance.

The previous algorithm of \cite{TomWelZam:2015} keeps a candidate phalanx
in the final ensemble if the phalanx is strong by itself
or strong in an ensemble with any other phalanx.
The old algorithm is vulnerable to including phalanxes that are strong individually but harmful to other phalanxes in the ensemble. 
Such weaknesses are avoided by the new criterion.

\subsection{Ensemble of phalanxes} Let $\hat{\pi}(\mathbf{x}_{(i)})$ be the probability
of homology vector for the test proteins obtained from 
LR applied to the phalanx $\mathbf{x}_{(i)}$.
The overall vector for the ensemble of phalanxes (EPX) is the average
\begin{equation} \label{eqn:epx:phalanx}
\hat{\pi}_{\text{EPX}} = \frac{\hat{\pi}(\mathbf{x}_{(1)}) + \hat{\pi}(\mathbf{x}_{(2)}) + \ldots + \hat{\pi}(\mathbf{x}_{(p)})}{p},
\end{equation}
which is used to compute assessment metrics.

\subsection{Ensemble of phalanxes across assessment metrics}

Let $\hat{\pi}_{\text{EPX},m}$ be probability of homology vector obtained from an ensemble of phalanxes optimized on metric $m$ of $M$ metrics. We exploit the diversity in assessment metrics to construct our proposed overall probability vector from an ensemble of phalanxes and metrics (EPX-EOM) by averaging:
\begin{equation} \label{eqn:epx:eom}
\hat{\pi}_{\text{EPX-EOM}} = \frac{\sum_{m = 1}^M \hat{\pi}_{\text{EPX},m}}{M}.
\end{equation}
Its ranking will be assessed in Section~\ref{sec:results} along with the rankings of EPX models for individual metrics.
Here, $M$---the number of diverse evaluation
metrics to ensemble---is application dependent. 
For the protein application, we consider APR and RKL, hence $M = 2$.  
TOP1 is evaluated in the next section but not used in the ensemble of metrics; as noted in Section \ref{sec:diversity:metrics}, APR provides better training discrimination of ranking performance at the beginning of the list.

\section{Results} \label{sec:results}

We first present summaries for ensembles fit to the protein homology training data.
EPX ensembles are built separately for the APR and RKL metrics and then combined in an overall ensemble of phalanxes across assessment metrics.
Their performances on the test data are then compared.

\subsection{EPX optimized for APR}\label{sec:res:apr}

Row 1 of Table \ref{tab:phalanxes:apr:rkl} summarizes the phalanxes formed using APR.
None of the 74 variables are filtered in the first stage of the algorithm,
hence all belong to one of the five candidate phalanxes found.
Two phalanxes are filtered out at the final stage, with the surviving three containing $36$, $23$ and $5$ variables, respectively, i.e., a total of $64$ variables.
\begin{table}
   \caption{\label{tab:phalanxes:apr:rkl}Number of candidate and filtered phalanxes identified by EPX and the sizes of the filtered phalanxes. The training data are used to optimize either APR or RKL.}
\centering
\fbox{%
\begin{tabular}{||c |c c |c ||}\hline\hline
\multirow{2}{*}{Metric} & \multicolumn{2}{|c|}{Number of phalanxes} & \\
\cline{2-3}
& Candidate & Filtered & Filtered-phalanx sizes \\\hline
APR & $5$ & $3$ & $36$, $23$, $5$ \\\hline
RKL & $7$ & $2$ & $29$, $27$ \\\hline\hline
\end{tabular}}
\end{table}

The second column of Table \ref{tab:epxapr:training:test} shows 10-fold CV APR for the three LR models fit to the three phalanxes and for their EPX-APR ensemble.  Here, CV is at the block level: the 153 blocks in the training data are randomly divided into 10 folds, where each fold contains the data from around 15 blocks. After refitting a phalanx's LR using 9/10 of the training data, estimated probabilities of homology are obtained for the remaining hold-out fold as usual in CV.  Repeating this process produces estimated probabilities for all the training data.  The three sets of estimated probabilities are then combined according to~(\ref{eqn:epx:phalanx}) for the EPX-APR ensemble.
The APR values for the three individual phalanxes are $0.794$, $0.785$ and $0.781$, respectively.
 

\begin{table}
\caption{\label{tab:epxapr:training:test}Training (10-fold CV) and test set APR for each of the three EPX phalanxes of variables found by maximizing APR and for the EPX-APR ensemble.
The top performance is highlighted in dark grey.}
\centering
\fbox{%
\begin{tabular}{||c|c|c||}\hline\hline
\multirow{2}{*}{Phalanx} & \multicolumn{2}{c||}{APR}\\\cline{2-3}
& Training (CV) & Test\\\hline
1 & 0.794 & 0.800\\ 
2 & 0.785 & 0.783\\ 
3 & 0.781 & 0.796\\\hline 
EPX-APR & \cellcolor{gray!77} 0.809 & \cellcolor{gray!77} 0.840\\ \hline\hline
\end{tabular}}
\end{table}

\begin{figure}
        \centering
        \subfloat[Homologous proteins]{
                \includegraphics[width=0.47\textwidth]{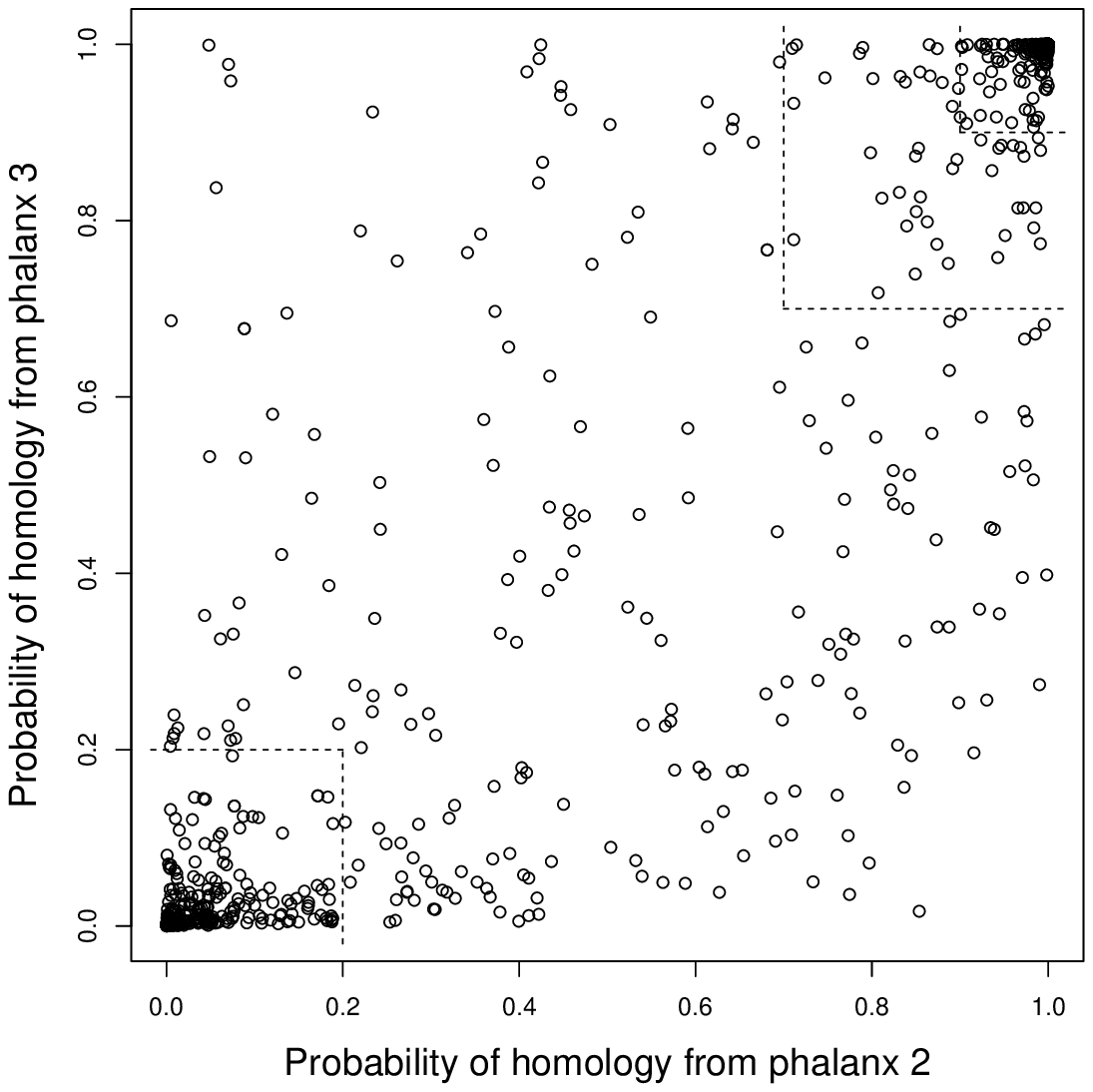}
                \label{HMPPX23}}
        \subfloat[Non-homologous proteins]{
                \includegraphics[width=0.47\textwidth]{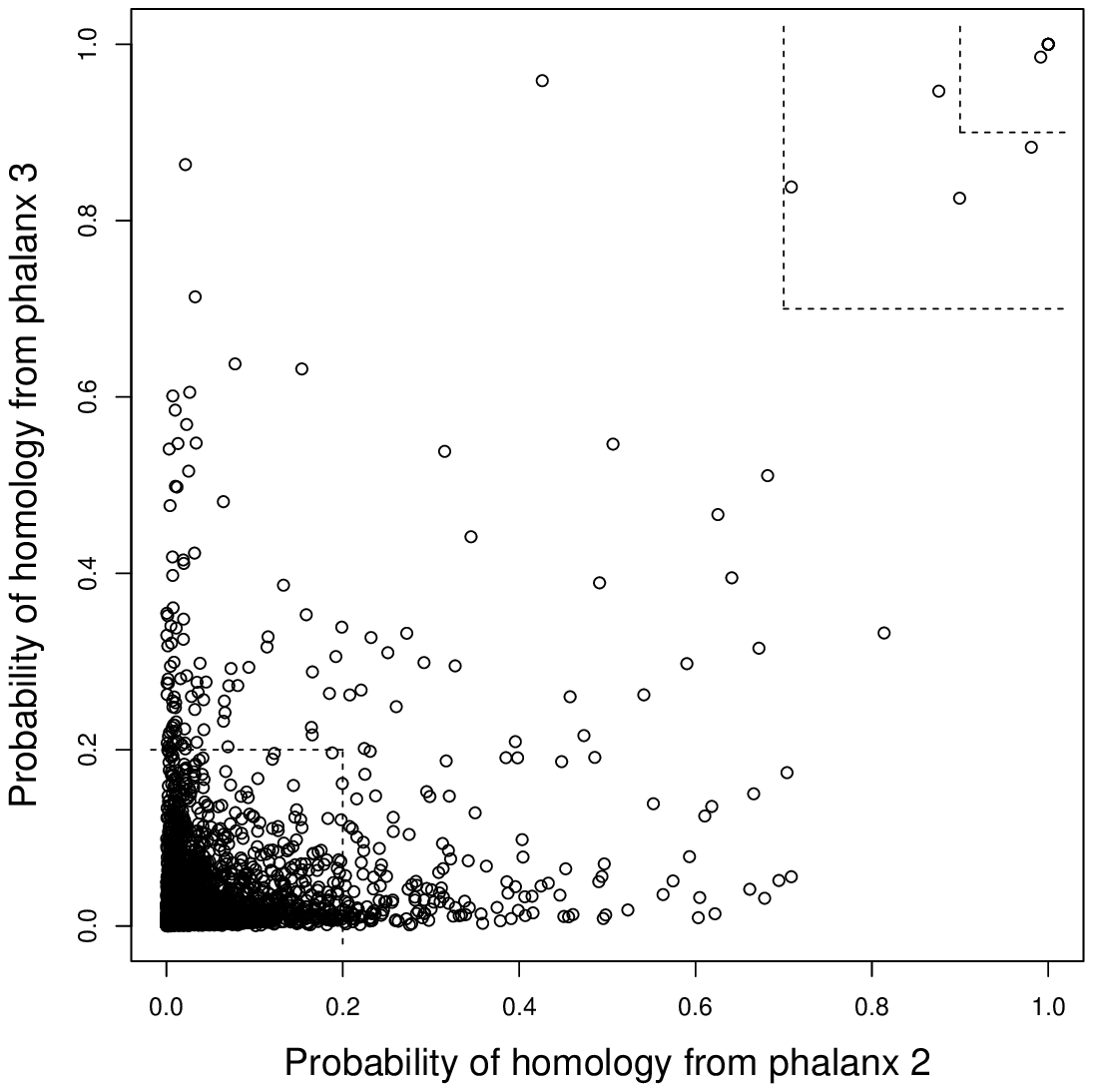}
                \label{NHMPPX23}}
\caption{Training data 10-fold CV LR probability of homology from phalanx 2 versus phalanx 3 for phalanx formation maximizing APR: (a) homologous proteins in the training data and (b) non-homologous proteins.}
        \label{fig:epxaprphalanxes123}
\end{figure}

Figure \ref{fig:epxaprphalanxes123} compares the CV probabilities of homology from phalanxes 2 and 3.  (These two phalanxes are chosen as we will argue they illustrate well diversity in predictions.)
Figure \ref{HMPPX23} focuses attention on the homologous proteins in the training data, 
where larger probabilities will lead to better ranking.
The smaller and larger top-right corners in Figure~\ref{HMPPX23}
contain $55.1\%$ and $58.8\%$, respectively, 
of all of the homologous proteins in the training data.
These true homologues will be ranked high overall, as there are very few proteins in the same high-probability regions for non-homologues in Figure \ref{NHMPPX23}.
It is also seen that some homologous proteins receive small probabilities.  The bottom-left corner of Figure \ref{HMPPX23} contains $24.4\%$ of all the homologous proteins in the
training data.  The same cell in Figure \ref{NHMPPX23} has far more proteins---homologous proteins account for only $0.22\%$ of cases where both probabilities are less than 0.2---but there is clearly potential for poor ranking of some homologues.
We call proteins that are difficult to predict {\em remote\/} homologues
because we suspect they are more separated from their native proteins than easy-to-predict {\em near\/} homologues.
Diversity of the probabilities from the two LR models helps for other homologues, however. 
Figure \ref{HMPPX23} contains homologous proteins in the top-left and bottom-right,
i.e., one of the two LRs assigns a high probability of homology,
and ensembling over the two LR models will pull these homologues ahead in the ranking.

Further diversity is present for phalanx $1$ versus phalanx $2$ and phalanx $1$ versus phalanx $3$,
and the second column of Table \ref{tab:epxapr:training:test}
shows that the EPX-APR ensemble across all three phalanx has an APR of $0.809$, an improvement over the performances of the three individual phalanxes.


Even better APR performance is found when the three LR models from the three phalanxes and
the EPX-APR ensemble are applied to the 150 blocks of test data.
True homology status is not revealed for the KDD Cup competition, but
submitting vectors of estimated probability of homology for the test proteins returns the results in
the last column of Table \ref{tab:epxapr:training:test}.
The APR values of $0.800$, $0.783$, and $0.796$ for the three phalanxes are about the
same or slightly better than from CV, 
while the $0.840$ for the EPX-APR ensemble is a noticeably improvement.

\subsection{EPX optimized for RKL} \label{sec:res:rkl}

Row 2 of Table \ref{tab:phalanxes:apr:rkl} summarizes the trained models when
phalanx formation minimizes RKL.
The algorithm filters none of the $74$ variables, which are then clustered into $7$ candidate phalanxes of which $2$ survive filtering.
A total of $56$ variables are used in phalanxes of $29$ and $27$ variables, respectively.

Results for 10-fold CV and for the test set are shown in  Table \ref{tab:epxrkl:training:test}.
Recall that RKL is a smaller-the-better metric.
While test performances for the two subsets and EPX-RKL are all worse than the estimates from CV,
it is seen that the EPX-RKL ensemble improves over the individual phalanxes for both the training and test sets.


\begin{table}
\caption{\label{tab:epxrkl:training:test}
Training (10-fold CV) and test set RKL for the two EPX phalanxes of variables found by minimizing RKL and for the EPX-RKL ensemble.
The top performance is highlighted in dark grey.}
\centering
\fbox{%
\begin{tabular}{||c|c|c||}\hline\hline
\multirow{2}{*}{Phalanx} & \multicolumn{2}{c||}{RKL}\\\cline{2-3}
& Training (CV) & Test\\\hline
1 & 59.2 & 67.6 \\ 
2 & 54.3 & 56.9 \\ \hline
EPX-RKL & \cellcolor{gray!77} 50.7 & \cellcolor{gray!77} 54.6 \\
\hline\hline 
\end{tabular}}
\end{table}

\begin{figure}
        \centering
        \subfloat[Homologous proteins]{
                \includegraphics[width=0.47\textwidth]{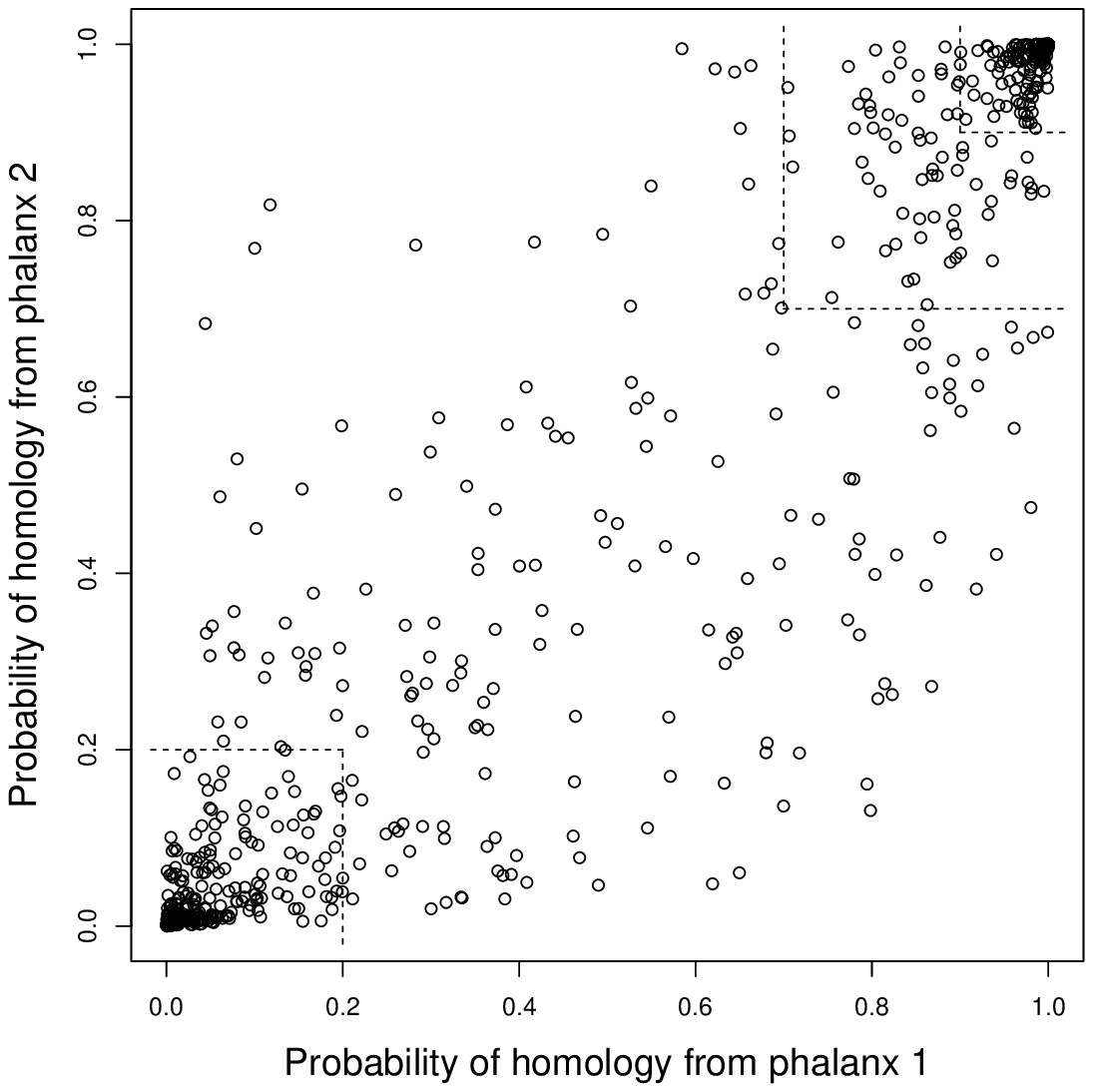}
                \label{HMPPX12}}
        \subfloat[Non-homologous proteins]{
                \includegraphics[width=0.47\textwidth]{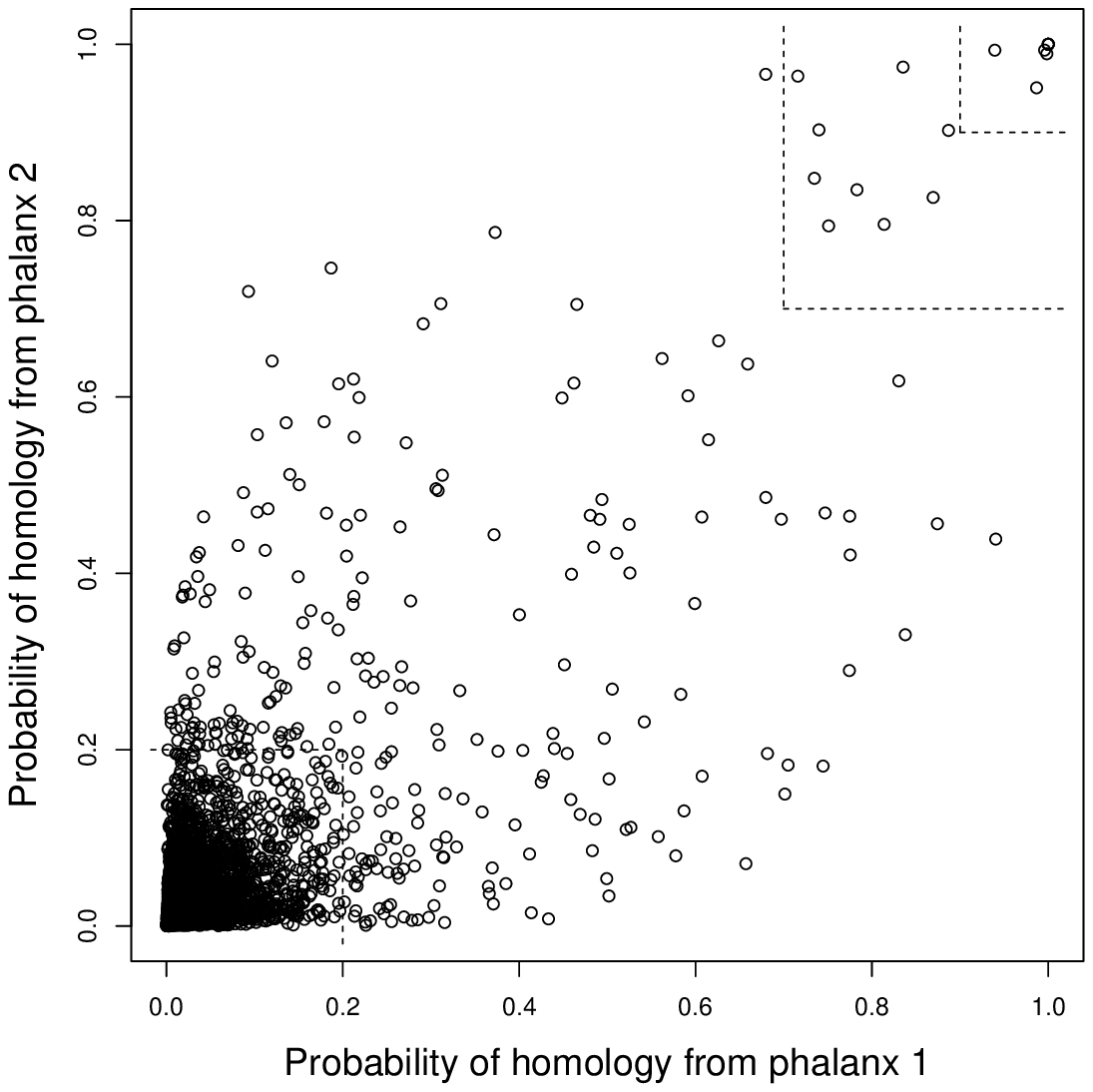}
                \label{NHMPPX12}}
\caption{Training data 10-fold CV LR probability of homology from phalanx 1 versus phalanx 2 for phalanx formation minimizing RKL: (a) homologous proteins in the training data and (b) non-homologous proteins.}
\label{fig:epxrklphalanxes12}
\end{figure}

Figure \ref{fig:epxrklphalanxes12} compares the CV probabilities of homology for the two phalanxes.
The smaller top-right and larger top-right corners of Figure \ref{HMPPX12} 
contain $60.0\%$ and $65.46\%$, respectively, of all of the homologous proteins in the training data,
better performance than seen in Figure \ref{HMPPX23} for APR.
Proteins in these regions will be ranked high, as the same regions in Figure~\ref{NHMPPX12}
for non-homologous proteins are sparse.
The lower-left corner of Figure \ref{HMPPX12} contains $20.2\%$ of the homologous proteins
but this is only $0.18\%$ of all proteins in this subregion across both Figure \ref{HMPPX12} and  Figure \ref{NHMPPX12},
slightly better than seen in Figure \ref{fig:epxaprphalanxes123}, and hence
EPX-RKL is slightly better in terms of harder to classify (i.e., remote) homologous proteins. 
Looking at the top-left and bottom-right corners of Figure \ref{HMPPX12} and Figure \ref{NHMPPX12} we see a modest number of proteins and hence some diversity in both panels,
which explains the superior performance of EPX-RKL in Table \ref{tab:epxrkl:training:test}
relative to models from either constituent phalanx.

\subsection{Ensemble of phalanxes across assessment metrics} \label{res:apr:rkl}

We next show that there is further diversity in predictions across the two metrics.
That diversity will be exploited by constructing an ensemble of phalanxes across assessment metrics called EPX-EOM:
\[
\ \hat{\pi}_{\text{EPX-EOM}} = \frac{\hat{\pi}_{\text{EPX-APR}} + \hat{\pi}_{\text{EPX-RKL}}}{2}.
\]



Columns $2$ to $4$ of Table \ref{tab:trainig:test:expaprrkl} show the $10$-fold CV training performances of EPX-APR, EPX-RKL and their ensemble EPX-EOM.
Recall that larger values of APR and TOP1 and smaller values of RKL are preferable.
Somewhat surprisingly, the APR performance of EPX-EOM, which averages classifiers built for another metric as well as APR, is superior to EPX-APR.
An analogous comment can be made for RKL.
Similarly TOP1 performance is superior for the aggregate.
Overall, then, for the training data the aggregated ensemble EPX-EOM improves on the performance of both ensembles constructed for a single objective uniformly across all metrics.

\begin{table}
\caption{\label{tab:trainig:test:expaprrkl}10-fold CV training performances and test performances - in terms of APR , TOP1 and RKL - of the ensemble of phalanxes
optimizing APR (EPX-APR), ensemble of phalanxes optimizing RKL (EPX-RKL), and their ensemble (EPX-EOM).
For each metric, the top performance is highlighted in dark grey.}
\centering
\fbox{%
\begin{tabular}{||l|ccc|ccc||}\hline\hline
\multirow{3}{*}{Ensemble} & \multicolumn{6}{c||}{Performance}\\\cline{2-7}
& \multicolumn{3}{c|}{Training (CV)} & \multicolumn{3}{c||}{Test}\\\cline{2-7}
& APR & TOP1 & RKL & APR & TOP1 & RKL\\\hline
EPX-APR & 0.809 & 0.843 & 62.3 & 0.840 & 0.920 & 59.0 \\ 
EPX-RKL & 0.798 & 0.824 & 50.7 & 0.835 & 0.900 & 54.6 \\ \hline
EPX-EOM & \cellcolor{gray!77} 0.816 & \cellcolor{gray!77} 0.850 & \cellcolor{gray!77} 50.1 & \cellcolor{gray!77} 0.844 & \cellcolor{gray!77} 0.920 & \cellcolor{gray!77} 54.1 \\ 
\hline\hline
\end{tabular}}
\end{table}



Table \ref{tab:wins:apr:rkl} compares the ranking performance of EPX-APR and EPX-RKL over the
training-data homologous compounds, by four categories. 
The categories are constructed based on the CV ranks from the aggregated ensemble EPX-EOM. 
To create categories of homologous compounds across blocks with differing sizes,
the EPX-EOM ranks are normalized to the range $[0, 1]$ within each block by 
dividing each rank by the number of candidates in the block.
These scores then divide the homologous proteins into four roughly equally size groups:
those with the smallest normalized ranks, in the range $[0.00000, 0.00264)$, are deemed ``very easy'' to classify, and so on up to $[0.02051, 0.50000)$ for ``very hard''.
For the ``very easy'' and ``easy'' categories, the ranks of EPX-APR are smaller (better) than those of EPX-RKL more often than they are larger.
For the ``hard'' and ``very hard'' categories, however, the comparison is reversed: 
EPX-RKL's ranks are more often smaller.
The suggestion here is that EPX-APR performs better for easier to rank close homologues,
whereas EPX-RKL performs better for harder to rank distant homologues.
In this sense the two classifiers are diverse.

\begin{table}
\caption{\label{tab:wins:apr:rkl}Comparison of ranking performance of EPX-APR versus EPX-RKL for all
training-data homologous compounds, categorized by the overall ranking performance of EPX-EOM.
The best classifier in each category is highlighted by dark grey.}
\centering
\fbox{%
\begin{tabular}{||llrrrr||}\hline\hline
EPX-EOM     &            & Homologous & \multicolumn{3}{c||}{Smaller rank from} \\
\cline{4-6}
normalized rank & Difficulty & proteins & APR & Tie & RKL\\ \hline
$[0.00000, 0.00264)$ & Very easy  & 323 & \cellcolor{gray!77}  59 & 223 &  41\\
$[0.00264, 0.00868)$ & Easy       & 325 & \cellcolor{gray!77} 117 & 103 & 105\\
$[0.00868, 0.02051)$ & Hard       & 323 &                     118 &  63 & \cellcolor{gray!77} 142 \\
$[0.02051, 0.50000)$ & Very hard  & 325 &                     103 &  34 &  \cellcolor{gray!77} 188\\
\hline
$[0.00000, 0.50000)$ &            & 1296 &                    397 & 423 & 476\\
\hline\hline 
\end{tabular}}
\end{table}

We next investigate further four example blocks chosen to illustrate how EPX-APR and EPX-RKL are complementary.
Table \ref{tab:MetricFourBlocks} gives the APR, TOP1 and RKL metrics from CV for training blocks 95, 216, 96 and 238.
In each block and for each metric the top-performer between EPX-APR and EPX-RKL is marked by dark grey
along with the EPX-EOM scores.
\begin{table}
\caption{\label{tab:MetricFourBlocks}Metrics APR, TOP1 and RKL for EPX-APR, EPX-RKL and EPX-EOM from CV for training blocks 95, 216, 96 and 238.
In each block, the top performance among EPX-APR and EPX-RKL is marked by dark grey.}
\centering
\fbox{%
\begin{tabular}{||rrr|lccr||}
  \hline\hline
      & \multicolumn{2}{c|}{Proteins} &        & \multicolumn{3}{c||}{Assessment Metrics}\\
      \cline{2-3} \cline{5-7}
Block & Total & Homologous           & Method & APR & TOP1 & RKL\\
\hline
\multirow{3}{*}{95} & \multirow{3}{*}{1120} &\multirow{3}{*}{1} & EPX-APR & \cellcolor{gray!77} 1.0000 & \cellcolor{gray!77} 1 & \cellcolor{gray!77} 1 \\ 
& & & EPX-RKL & 0.0909 & 0 & 11 \\ 
& & & EPX-EOM & 1.0000 & 1 & 1 \\\hline 
\multirow{3}{*}{216} & \multirow{3}{*}{1068} & \multirow{3}{*}{3} & EPX-APR & \cellcolor{gray!77} 0.9167 & \cellcolor{gray!77} 1 & \cellcolor{gray!77} 4 \\ 
& & & EPX-RKL & 0.8333 & 1 & 6 \\ 
& & & EPX-EOM & 0.9167 & 1 & 4 \\ \hline
\multirow{3}{*}{96} & \multirow{3}{*}{974} & \multirow{3}{*}{2} & EPX-APR & 0.1270 & 0 & 492 \\ 
& & & EPX-RKL & \cellcolor{gray!77} 0.1275 & \cellcolor{gray!77} 0 & \cellcolor{gray!77} 394 \\ 
& & & EPX-EOM & 0.1274 & 0 & 415 \\ \hline
\multirow{3}{*}{238} & \multirow{3}{*}{861} & \multirow{3}{*}{50} & EPX-APR & 0.5651 & 1 & 584 \\ 
& & & EPX-RKL & \cellcolor{gray!77} 0.7811 & \cellcolor{gray!77} 1 & \cellcolor{gray!77} 249 \\ 
& & & EPX-EOM & 0.7125 & 1 & 346 \\ 
   \hline\hline
\end{tabular}}
\end{table}
Blocks $95$ and $216$ are fairly easy as both EPX-APR and EPX-RKL assign relatively small ranks to all their respective homologues,
i.e., the homologues are close to their respective native proteins.
Nonethess, EPX-APR does better, placing the single homologue in Block 95 at the very beginning of the list and placing the three homologues in Block 216 in positions 1, 2, and 4.
APR is sensitive to small changes in ranks of homologous proteins at the head of a list. 
EPX-RKL performs slightly worse but has no impact on the aggregate performance: EPX-EOM's metrics
are exactly the same as EPX-APR's.
Block 96 is apparently much harder. EPX-APR and EPX-RKL both place a homologue in the fourth position,
but the second---a distant homologue---is far down the list at position 492 or 394, respectively.
It is seen, however, that EPX-RKL performs better for both APR and RKL and that it's superior performance
is almost matched by EPX-EOM.
Block 238 is also difficult in the sense that there are 50 homologues to find, 
and the last homologue is always well down the list.
Again, EPX-RKL performs better, even for the APR metric, and EPX-EOM's metrics are closer to those of EPX-RKL. 
It appears that EPX-APR performs better for blocks with close homologous proteins that are easy to rank, whereas EPX-RKL performs better for distant homologues.

Table \ref{tab:trainig:test:expaprrkl} confirms using the test data that EPX-EOM combines its constituent classifiers to advantage.  It is  superior or as good as either EPX-APR or EPX-RKL
uniformly across the metrics,
mirroring what was observed for the training data.

Finally, the test metrics are compared to the results for the winners
of the $2004$ knowledge discovery and data mining cup competition.
Four different research groups (Weka, ICT.AC.CN, MEDai/Al Insight and PG445 UniDo) were declared winners
of the protein homology section, with Weka the overall winner.

The top part of Table \ref{tab:CompKddEpx} summarizes the test metrics for
EPX-APR, EPX-RKL and EPX-EOM, while the central part shows results for the winners of the competition.
As a further benchmark, the lower part of Table \ref{tab:CompKddEpx} presents results from RF \citep{Breiman:2001}
and an ensemble denoted EPX-RF \citep{TomWelZam:2015}, 
where the underlying classifier is RF instead of logistic regression and EPX is optimized for APR 
(equivalent to the AHR employed by \cite{TomWelZam:2015}.).
\begin{table}
\caption{\label{tab:CompKddEpx}Test metrics for EPX-APR, EPX-RKL and EPX-EOM (top),
the winners of the $2004$ knowledge discovery and data mining (KDD) cup competition (centre),
and RF and EPX-RF (bottom). The best value of each metric is highlighted in dark grey.}
\centering
\fbox{%
\begin{tabular}{||l|rc|rc|rc|c||}
  \hline \hline
Methods & APR & Rank & TOP1 & Rank & RKL & Rank & Average Rank\\\hline
EPX-APR          & 0.8398 & 4 & \cellcolor{gray!77} 0.9200 & 2   & 59.027  & 7 & 4.333\\
EPX-RKL          & 0.8349 & 6 & 0.9000 & 6   & 54.567  & 6 & 6.000\\
EPX-EOM     & \cellcolor{gray!77} 0.8437 & 1 & \cellcolor{gray!77} 0.9200 & 2   & 54.080  & 4 & 2.333\\\hline\hline
Weka             & 0.8409 & 3 & 0.9067 & 5   & 52.447  & 2 & 3.333\\
ICT.AC.CN        & 0.8412 & 2 & 0.9133 & 4   & 54.087  & 5 & 3.667\\
MEDai/Al Insight & 0.8380 & 5 & \cellcolor{gray!77} 0.9200 & 2   & 53.960  & 3 & 3.333\\
PG445 UniDo      & 0.8300 & 7 & 0.8667 & 9   & \cellcolor{gray!77} 45.620  & 1 & 5.667\\\hline\hline
RF               & 0.8089 & 9 & 0.8733 & 7.5 & 143.733 & 9 & 8.500\\
EPX-RF           & 0.8140 & 8 & 0.8733 & 7.5 & 82.307  & 8 & 7.833\\ \hline\hline
\end{tabular}}
\end{table}
The methods are also ranked in the table by metric and overall. 
For APR, EPX-EOM is ranked first followed by ICT.AC.CN, and so on. 
Three methods, EPX-EOM, EPX-APR and MEDai/Al Insight, tie in first place for TOP1 and receive an average rank of $2$. 
PG445 UniDo is ahead of all other methods in terms of RKL but performs fairly poorly for APR and TOP1.
To give an overall result, the final column of Table~\ref{tab:CompKddEpx} averages the ranks across the three metrics.
EPX-EOM has the best average rank at $2.333$, with Weka and MEDai/Al Insight
tying next with $3.333$. 

Although the competition is over, new predictions can be submitted to the $2004$ KDD Cup website. No new results---as of July 30, 2019---are better or even equal to our results in terms of
APR ($0.8437$) and TOP1 ($0.92$).

\subsection{Computational gain of EPX-LR over EPX-RF}

Table \ref{tab:CompTime} shows the computation times (training and validation) for EPX-LR, EPX-RF and RF. The table also shows the number of processors used and amount of memory allocated for parallel execution of the three ensembles running on the machine bugaboo\footnotemark[1] of the Western Canada Research Grid (WestGrid).
For each of the $32$ processors $8$ GB of memory was allocated. 
The computation times to train and validate
EPX-RF and EPX-LR were 1569 and 70--80 minutes, respectively.
(EPX-LR is run with several metrics, hence the range of times.)
The time for EPX-LR based on logistic regression 
is much smaller than for EPX-RF with RF as the base classifier.
Through parallel computation, the computing time of EPX-LR was brought down to be similar to a
single RF, for which training and validation 
using one processor and $8$ GB memory took $67$ minutes.
One can also easily assign more processors to further reduce the computation time of EPX.
\footnotetext[1]{\url{http://www.westgrid.ca/support/quickstart/bugaboo}: Accessed November 06, 2016.}
\begin{table}
\caption{\label{tab:CompTime}Computation time in minutes for EPX-LR, EPX-RF and RF using Bugaboo of WestGrid}
\centering
\begin{tabular}{||l|ccc||}
  \hline \hline
\multirow{2}{*}{Ensembles} & Number of & Memory allocation & Elapsed time\\
& Processors & (GB) & (Minute)\\\hline
RF & $1$ & $8$ & $67$\\
EPX-RF & $32$ & $8$ & $1569$\\
EPX-LR & $32$ & $8$ & $70$-$80$\\\hline\hline
\end{tabular}
\end{table}

\section{Summary and conclusion} \label{dis:con}

An ensemble is an aggregated collection of models. To build a powerful ensemble,
the constituent models need to be strong and
diverse. In this article, ``strong'' stands for good predictive ranking ability
of a model. The EPX algorithm groups
a number of variables into various phalanxes guided by a ranking objective. Two phalanxes and their models
are diverse in the sense that they predict well different sets of homologous
proteins. The algorithm keeps grouping
variables into phalanxes until a set of variables help each other in a model, and
generally ends up with more than one phalanx. 
In a nutshell, EPX seeks to place good variables together in a
phalanx to increase its strength, while exploiting different phalanxes to induce diversity between models.

A nice property of EPX is filtering
weak/noise variables. A variable is considered weak if it shows poor
predictive performance: $(a)$ individually in a model, $(b)$ jointly when paired with another variable
in a model, and $(c)$ jointly when ensembled with another variable in different models.
The algorithm also filters harmful phalanxes of variables.
Harmful stands for weakening the predictive ranking of an ensemble
compared with the best predictive ranking of individual phalanxes.
A candidate phalanx is filtered out if it appears harmful
with any other phalanx used in the final ensemble.
Thus, the merging/clustering step of the algorithm of phalanx formation
is sandwiched by filtering weak variables and harmful
candidate phalanxes.

A previously less highlighted property of our EPX algorithm is
model selection. Regular model selection methods---such as forward stepwise selection,
backward elimination, ridge regression \citep{HastiTibFried:2009}, 
and lasso \citep{Tibshirani:1996}---select one subset (phalanx) of variables and build one model.
In contrast, EPX selects a collection
of models and aggregates these in an ensemble. Each of
these models can be considered as an alternative to the
models conveniently selected by other methods. Furthermore,
the constituent models in EPX predict diverse collections of homologous proteins here, which
is similar to examining a problem from different vantage points.

Previous EPX applications have used RF as the underlying classifier.
Here, though, LR is a better base learner.
The results of EPX using LR are reasonably close to
the winners of the $2004$ KDD Cup competition. These findings are encouraging and suggest
considering suitable base learners such as recursive partitioning \citep{BreimanTree:1984}, neural networks
\citep{Ripley:1996},
etc., as best for a given application.

EPX is independently optimised for average precision
and rank last here. This exemplifies the flexibility of EPX. The underlying base classifier need not be rewritten, rather a specific metric is applied to guide phalanx formation.  The choice of assessment metrics can be 
application-dependent, with the goal finding an ensemble of metrics with good overall performance.
As we have seen here, EOM can provide uniformly better performance even against models tuned specifically for one criterion.
APR is useful to rank close homologues, while RKL is helpful to rank distant homologues. 
Future work includes 
extending EPX-EOM to linear regression and multi-class classification problems using 
variants of metrics such as mean squared error and misclassification rate, etc.
It would also be interesting to examine if improvements can be achieved by aggregating ensembles of phalanxes based on
different base learners. 

\section*{Acknowledgements}
We thank Professor Ron Elber, University of Texas at Austin, Texas, USA, for his help
to understand the feature variables of the protein homology data.
We acknowledge support for this project
from the Natural Sciences and Engineering Research Council of Canada
(NSERC grants RGPIN-2014-04962 and RGPIN-2014-05227).
The WestGrid computational facilities of Compute Canada are also gratefully acknowledged.

\bibliographystyle{rss}
\bibliography{bib_JRSSC}

\section*{Appendix}

\subsection*{Algorithm 1: Phalanx formation by optimizing average precision}

\begin{description}
\item[1. Determine the reference distribution.] Let $a$ be the metric average precision. Use random permutation to determine the reference distribution of $a$. To do that, record the orders of the candidate proteins, randomly permute the response variable $y$, compute the metric $a$ within each block, and then average the computed $a$ across the blocks. Repeat the process as many times as required (we used $2000$). Record $\alpha$th quantile $a_{\alpha}$ and median $a_{0.50}$ of the reference distribution of $a$.
Use $\alpha = 0.95$ when optimizing average precision.

\item[2. Save the CV indices.] Generate a random set of indices for   $10$-fold CV defined at the block level. Use the CV indices to evaluate every models during phalanx formation.

\item[3. Filter weak variables.] Let $\xVec = \{x_1, x_2, \cdots, x_d\}$ be a vector of $d$ variables. For each variable $x_i$ $(i = 1, 2, \cdots, d)$, 
fit an LR model and obtain the CV probability of homology vector $\hat{\pi}(x_i)$. 
Calculate average precision $a_i = a\left(\hat{\pi}(x_i)\right)$ which represents the strength of the $i$th variable. The larger the $a_i$,
the stronger the variable.

Fit an LR model using $x_i \cup x_j$ ($j > i = 1, 2, \cdots, d$) and obtain CV probability of homology vector $\hat{\pi}(x_i \cup x_j)$.
Calculate $a_{ij} = a(\hat{\pi}(x_i \cup x_j))$ which represents the joint strength of $x_i$ and $x_j$ when putting them together in a model.
The larger the values of $a_{0.50} + a_{ij} - a_{j}$, the stronger the marginal strength of $x_i$ having $x_j$ in the same model.

Calculate $a_{\overline{ij}} = a((\hat{\pi}(x_i) + \hat{\pi}(x_j))/2)$ which represents the strength of $x_i$ when ensembling with another variable $x_j$ $(j > i = 1, 2, \cdots, d)$. The larger the values of
$a_{0.50} + a_{\overline{ij}} - a_{j}$, the stronger the marginal strength of $x_i$ having $x_j$ in a different model.

Combine all of the three aspects and filter the $i$th variable $x_i$ out if
\[
\ \max[a_i, a_{0.50} + a_{ij} - a_{j}, a_{0.50} + a_{\overline{ij}} - a_{j}] < a_{\alpha} \ ; \ \forall j (> i) = 1, \ldots, d,
\]
where $\alpha = 0.95$. Pass a set of useful variables $\{x_1, \ldots, x_s\}, s \leq d$ to the next step \emph{merge variables into phalanxes}.

\item[4. Merge variables into phalanxes.] Assign $x_i$ to $\mathbf{g}_i$, where
$\mathbf{g}_i$ is considered as a set of variables. In order to merge the set of variables $\mathbf{g}_i$ into phalanxes, minimize the following criterion
\[
\ m_{ij} = \frac{\max(a_{\overline{ij}}, a_i, a_j)}{a_{ij}}
\]
over all possible pairs $(i, j); j > i = 1, \ldots, s$. 
If $m_{ij} < 1$, merge the sets $\mathbf{g}_i$ and $\mathbf{g}_j$ together. That is, merge the two old sets to form a new set. After each merge, decrease the number of sets $s$ by $1$. Continue merging until
$m_{ij} \geq 1$ for all $i, j$ which suggests that merging either degrades individual performances or their ensembling performance.
Assign $s \rightarrow c$ and $\mathbf{g}_i \rightarrow \mathbf{x}_i$ and pass
the set of candidate phalanxes $\left\{\{\mathbf{x}_1\}, \{\mathbf{x}_2\}, \ldots, \{\mathbf{x}_c\}\right\}$ to the \emph{filter candidate phalanxes} step.

\item[5. Filter candidate phalanxes.] To detect weak and/or harmful phalanxes and thus to filter, minimize the
following criterion
\[
\ f_{ij} = \frac{a_{\overline{ij}}}{\max(a_i, a_j)}
\]
over all possible pairs of candidate phalanxes $(i, j); j > i = 1, \ldots, c$.
If $f_{ij} \leq 1$, filter the weaker phalanx:
If $a_i \leq a_j$, filter out the $i$th phalanx; otherwise, filter out the $j$th phalanx.
After filtering a phalanx, decrease the number of candidate phalanxes $c$ by $1$.
Hierarchically filter the phalanxes until $f_{ij} > 1$ or $c = 1$. 
Finally, deliver a set of phalanxes $\left\{\{\mathbf{x}_{(1)}\}, \{\mathbf{x}_{(2)}\right\}, \ldots, \{\mathbf{x}_{(p)}\}\}$.
\end{description}

\subsection*{Algorithm 2: Phalanx formation by optimizing rank last}

\begin{description}
\item[1. Determine the reference distribution.] Let $r$ be the metric rank last. Similar to algorithm 1, use random permutation to determine the reference distribution of $r$. Record $\alpha$th quantile $r_{\alpha}$ and median $r_{0.50}$ of the reference distribution of $r$. Use $\alpha = 0.05$ when optimizing rank last.

\item[2. Save the CV indices.] Same as algorithm 1.

\item[3. Filter weak variables.] Proceed similarly as in algorithm 1 and filter the $i$th variable $x_i$ out if
\[
\ \min[r_i, r_{0.50} + r_{ij} - r_{j}, r_{0.50} + r_{\overline{ij}} - r_{j}] > r_{\alpha} \ ; \ \forall j (> i) = 1, \ldots, d,
\]
where $\alpha = 0.05$. Pass a set of useful variables $\{x_1, \ldots, x_s\}, s \leq d$ to the next step \emph{merge variables into phalanxes}.

\item[4. Merge variables into phalanxes.] Assign $x_i$ to $\mathbf{g}_i$. To merge $\mathbf{g}_i$ into phalanxes, maximize the following criterion
\[
\ m_{ij} = \frac{\min(r_{\overline{ij}}, r_i, r_j)}{r_{ij}}
\]
over all possible pairs $(i, j); j > i = 1, \ldots, s$. 
If $m_{ij} > 1$, merge the sets $\mathbf{g}_i$ and $\mathbf{g}_j$ together. After each merge, decrease the number of sets $s$ by $1$. Continue merging until
$m_{ij} \leq 1$ $\forall i, j$.
Assign $s \rightarrow c$ and $\mathbf{g}_i \rightarrow \mathbf{x}_i$ and pass
the vector of candidate phalanxes $\left\{\{\mathbf{x}_1\}, \{\mathbf{x}_2\}, \ldots, \{\mathbf{x}_c\}\right\}$ to the \emph{filter candidate phalanxes} step.

\item[5. Filter candidate phalanxes.] To filter weak and/or harmful phalanxes, maximize the
following criterion
\[
\ f_{ij} = \frac{r_{\overline{ij}}}{\min(r_i, r_j)}
\]
over all possible pairs of candidate phalanxes $(i, j); j > i = 1, \ldots, c$.
If $f_{ij} \geq 1$, filter the weaker phalanx:
If $r_i \geq r_j$, filter out the $i$th phalanx; otherwise, filter out the $j$th phalanx.
After filtering a phalanx, decrease the number of candidate phalanxes $c$ by $1$.
Hierarchically filter the phalanxes until $f_{ij} < 1$ or $c = 1$. 
Finally, deliver a set of phalanxes $\left\{\{\mathbf{x}_{(1)}\}, \{\mathbf{x}_{(2)}\right\}, \ldots, \{\mathbf{x}_{(p)}\}\}$.
\end{description}

\end{document}